\documentclass{article} 
\usepackage{iclr2023_conference,times}

\usepackage{amsmath,amsfonts,bm}

\def\eqref#1{equation~\ref{#1}}

\def\1{\bm{1}}

\DeclareMathAlphabet{\mathsfit}{\encodingdefault}{\sfdefault}{m}{sl}
\SetMathAlphabet{\mathsfit}{bold}{\encodingdefault}{\sfdefault}{bx}{n}

\usepackage{hyperref}
\usepackage{url}

\usepackage[utf8]{inputenc} 
\usepackage[T1]{fontenc}    
\usepackage{hyperref}       
\usepackage{url}            
\usepackage{booktabs}       
\usepackage{amsfonts}       
\usepackage{nicefrac}       
\usepackage{microtype}      
\usepackage{xcolor}         
\usepackage{amsmath}
\usepackage{graphicx}

\title{Label Calibration for Semantic Segmentation\\ Under Domain Shift}

\author{Ondrej Bohdal$^{1}$\thanks{Work done during an internship at Samsung AI Center, Cambridge.}, Da Li$^{2}$, Timothy Hospedales$^{1,2}$ \\
$^{1}$The University of Edinburgh \ $^{2}$Samsung AI Center, Cambridge \\
 $^{1}$\texttt{\{ondrej.bohdal,t.hospedales\}@ed.ac.uk} \ $^{2}$\texttt{da.li1@samsung.com}
}

\iclrfinalcopy 
\begin{document}

\maketitle

\begin{abstract}
Performance of a pre-trained semantic segmentation model is likely to substantially decrease on data from a new domain. We show a pre-trained model can be adapted to unlabelled target domain data by calculating soft-label prototypes under the domain shift and making predictions according to the prototype closest to the vector with predicted class probabilities. The proposed adaptation procedure is fast, comes almost for free in terms of computational resources and leads to considerable performance improvements. We demonstrate the benefits of such label calibration on the highly-practical synthetic-to-real semantic segmentation problem.
\end{abstract}

\section{Introduction}
Domain shift represents a significant challenge when deploying models to real-world problems \citep{kouw2021daReview,csurka2022visual}. When the target data distribution does not match the training data distribution, the performance of the model will suffer, which can be a safety-critical issue. For example, autonomous vehicles \citep{Bojarski2016EndCars} are likely to operate under many different conditions and hence are likely to encounter domain shifts leading to decreases in performance.
Perhaps more fundamentally, the model navigating the autonomous vehicle may have been trained on synthetic data as it is easier to obtain a large quantity of such data with labels. This leads to the synthetic-to-real shift during deployment, where the model needs to be adapted to real data.

We focus on the problem setting where we need to adapt a pre-trained model to a new unlabelled dataset or domain. The model was pre-trained on source data, but these are no longer available during adaptation (unsupervised source-free domain adaptation \citep{Liang2020DoAdaptation}). This set-up has recently attracted significant attention \citep{Liang2020DoAdaptation, Huang2021ModelData, Kundu2020UniversalAdaptation}, first within image classification but later also within semantic segmentation \citep{Kundu2021GeneralizeSegmentation} and other areas of computer vision \citep{Li2020ModelData}. SFDA is practical because it does not require access to the source dataset and can also be significantly faster than training an adapted model from scratch.

However, most existing SFDA algorithms are too slow to update on an autonomous embedded platform, because they use back-propagation and multiple passes over the target dataset. We are interested in exploring to what extent adaptation can still be performed without back-propagation in a single pass over the dataset. If this is possible, it will be significantly more practical to use adaptation in deployed applications, potentially even in an online or streaming mode.

Our key contribution is a fast method that adapts a pre-trained semantic segmentation model to a new unlabelled dataset. The method is entirely feed-forward, does not require access to the model parameters (black-box SFDA) and can be viewed as calibration of labels under domain shift. We find that soft labels obtained by a pre-trained model under domain shift act as useful prototypes for making predictions. We empirically demonstrate that predicting the class according to the nearest soft-label prototype leads to improved performance in the presence of domain shift.

\section{Related work}

\subsection{Source-free domain adaptation}
Standard unsupervised domain adaptation assumes access to both labelled source domain data and unlabelled target domain data \citep{Ganin2016Domain-AdversarialNetworks, Sun2016DeepAdaptation,Saito2018MaximumAdaptation}. However, the need to access source data during adaptation has been challenged by \citet{Liang2020DoAdaptation}, who have shown that strong performance can be obtained even without access to the source data. In such cases a pre-trained model is fine-tuned to unlabelled target domain data. For example, a model can be fine-tuned using unlabelled data by maximizing information transferred from the source model or by using self-supervised pseudo-labelling \citep{Liang2020DoAdaptation}. Many alternative methods have been proposed, including historical contrastive learning \citep{Huang2021ModelData} and Universal SFDA (USFDA) \citep{Kundu2020UniversalAdaptation}. It has also been shown that updating the batch normalization (BN) statistics on unlabelled target domain data can also significantly improve the performance of the model on the target domain \citep{Schneider2020ImprovingAdaptation,Zhang2021AdaptiveShift}.

\subsection{Source-free domain adaptation for semantic segmentation}
Approaches for source-free domain adaptation have recently been developed also in the context of semantic segmentation \citep{Kundu2021GeneralizeSegmentation}. The method from  \citet{Kundu2021GeneralizeSegmentation} consists of two main steps: 1) vendor-side preparation and 2) client-side preparation. Vendor-side preparation consists of training on synthetic source data with a large variety of strong augmentations such as weather augmentation \citep{imgaug,Michaelis2019BenchmarkingComing} and style augmentation \citep{Jackson2019StyleRandomization}. Such ERM training has also been shown to be a highly competitive domain generalization approach that is hard to beat  \citep{Gulrajani2020InGeneralization}. Client-side preparation uses a multi-head framework that tries to extract reliable target pseudolabels for self-training.

\section{Methods}
We first pre-train a model on source data following  \citet{Kundu2021GeneralizeSegmentation}, so we pre-train the model on synthetic data with a large number of data augmentations. We use the pre-trained model to segment target domain (real-world) images and predict the probabilities of different classes for each pixel. To adapt the model, we construct soft-label prototypes that are then used to calibrate the predictions on new unseen target domain data.

\begin{figure}[h!]
  \begin{center}
  \includegraphics[width=0.8\columnwidth]{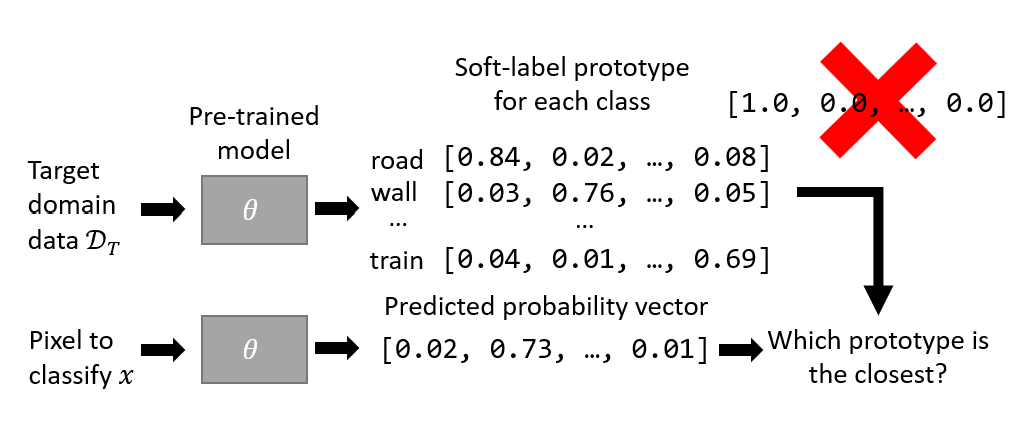}
  \caption{Illustration of our method. Target domain data $\mathcal{D}_T$ are used to construct a soft-label prototype for each class under domain shift (each prototype is a valid probability distribution). When we try to classify a pixel $x$ to perform semantic segmentation, we use the pre-trained model to predict a vector of probabilities of the different classes. Rather than taking the most likely class, we make a prediction by finding the closest soft-label prototype to the predicted probability vector.}
  \label{fig:illustration}
  \end{center}
\end{figure}

We make predictions for each pixel by predicting the class of the soft-label prototype closest to the predicted probability vector, using Euclidean distance. We illustrate the details of our method in Figure \ref{fig:illustration}. The key intuition for our method is that due to the distortions caused by domain shift, better predictions can be made by finding the closest probability profile under domain shift, instead of simply selecting the class with the highest probability.

Each prototype is a vector of $C$ elements, where $C$ is the number of predicted classes. All prototypes together are represented as a $C\times C$ matrix. The prototype of each class is constructed by taking a confidence-weighted average of the predicted soft labels across all pixels predicted to be of the given class. We use pixels from all examples in the training part of the unlabeled target dataset.

Let us define the prototypes formally using a formula. We denote the probability of class $c$ (scalar) for example $n$ for pixel located at $w, h$ as $p_{n,c,w,h}$ –- this is the output of the pre-trained model. The whole probability vector for different classes is denoted $\boldsymbol{p}_{n, . ,w,h}$ and has $C$ elements. $1\left[p_{n,c,w,h}==\max(p_{n, . ,w,h})\right]$ will be 1 if $c$ is the most probable class of pixel located at $w, h$ of example $n$ –- and will be 0 otherwise. We define $$m_{n,c,w,h}=1\left[p_{n,c,w,h}==max(p_{n, . ,w,h})\right]p_{n,c,w,h}$$ as a confidence-weighted indicator saying if the most probable class of the given pixel is $c$. We then calculate the prototype of class $c$ (vector) as follows:
$$\boldsymbol{\mu}_c=\frac{\sum_{n,w,h} \boldsymbol{p}_{n,.,w,h} m_{n,c,w,h}}{\sum_{n,w,h} m_{n,c,w,h}}$$

Each prediction is associated with uncertainty reflecting confidence of the model about the prediction. We use the confidence weighting because when the model is more confident about its prediction, it is more likely that the corresponding soft-label vector will be more useful for constructing the overall label prototype of the given class. Each prototype is a probability vector for the given class.

\section{Experiments}
\subsection{Set-up}
We implement the pre-training by directly using the pre-trained models provided by \citet{Kundu2021GeneralizeSegmentation}, so we only focus on the adaptation part. We also use the code provided by \citet{Kundu2021GeneralizeSegmentation} to implement our approach and compare with their approach. We show the benefits of our approach on the highly practical synthetic-to-real semantic segmentation problem. We use GTA5 \citep{Richter2016PlayingGames} and Synthia datasets \citep{Ros2016TheScenes} as the synthetic source domains and Cityscapes \citep{Cordts2016TheUnderstanding} as the real-world target domain. We perform experiments using DeepLabv2 \citep{Chen2017DeepLab:CRFs} with ResNet101 \citep{He2015DeepRecognition} backbone.

We consider three main baselines: 1) directly using the strong ERM pre-trained model, 2) updating batch normalization (BN) statistics, 3) shallow self-training inspired by \citep{Kundu2021GeneralizeSegmentation}. We also show the performance of our own runs of the method from \citet{Kundu2021GeneralizeSegmentation} -- \textit{GtA (Generalize then Adapt)}. We are interested in understanding how well we can adapt the model under the limited compute resources. \textit{GtA} gives us context about what an upper bound of the performance could be -- \textit{GtA} is an example of a standard SFDA method that uses back-propagation and performs many passes over the dataset. In this sense we can view our approach as a fast re-calibration of the predictions under domain shift rather than a full-scale source-free domain adaptation method. This view also reflects the magnitude of improvements associated with our proposed method.

We use minibatch size of 4 and there are 19 classes. When constructing the prototypes we do a single pass on the training part of the Cityscapes dataset. The evaluation is done on a separate test part of Cityscapes. In case some classes are never predicted on the training part of Cityscapes, we use standard one-hot vectors as soft-label prototypes for these classes.

In the case of the BN update baseline, we do one pass over the training part of the target domain and update the BN statistics for all layers of the model. Our shallow self-training baseline only updates the top layer and does a single pass over the training part of Cityscapes (same as us). We train the \textit{GtA} method from \citet{Kundu2021GeneralizeSegmentation} for 50,000 iterations (default value in \citep{Kundu2021GeneralizeSegmentation}).

\subsection{Results}
We show the results of our experiments in Table \ref{tab:sfda}. Our approach brings a non-negligible improvement in mIoU of around 1.0\% for both GTA5 and Synthia datasets compared to the strong pre-trained ERM baseline. The improvement comes almost for free in terms of computational costs as we analyse in Section \ref{sec:analysis}. Even though \textit{GtA} gives a relatively strong improvement if trained fully, it gives only a marginal improvement compared to our approach if given a similar amount of time, as we will also see in Section \ref{sec:analysis}. BN updates hurt performance in this case, likely because the pre-training was done with a variety of strong data augmentations. Shallow self-training can improve the performance marginally, but its performance has been inconsistent.

\begin{table}[h]
  \caption{Our Label calibration leads to considerable improvements at only a small computational cost. Mean mIoU (\%) and standard deviation across three runs. mIoU is reported across 19 classes for GTA5 $\rightarrow$ Cityscapes, and across 16 and 13 classes for Synthia $\rightarrow$ Cityscapes (in line with standard conventions). Note that \textit{GtA} is reported to give us context about the performance of standard resource-intensive SFDA methods i.e. to give us an indication of upper-bound performance.}
  \label{tab:sfda}
  \begin{center}
  \begin{tabular}{lcc}
    \toprule
    Approach     & GTA5 $\rightarrow$ Cityscapes & Synthia $\rightarrow$ Cityscapes \\
    \midrule
    Pre-trained model & 43.33 $\pm$ 0.00 & 40.25 $\pm$ 0.00 / 46.72 $\pm$ 0.00 \\
    BN updates & 41.14 $\pm$ 0.49 & 38.43 $\pm$ 0.05 / 44.22 $\pm$ 0.07 \\
    Shallow self-training & 44.01 $\pm$ 0.31 & 40.15 $\pm$ 0.13 / 46.68 $\pm$ 0.19 \\
    Label calibration (ours) & 44.36 $\pm$ 0.01 & 41.54 $\pm$ 0.02 / 48.03 $\pm$ 0.02 \\
    \midrule
    GtA \citep{Kundu2021GeneralizeSegmentation} & 46.98 $\pm$ 0.16 & 42.21 $\pm$ 3.48 / 48.91 $\pm$ 3.74\\
    \bottomrule
  \end{tabular}
  \end{center}
\end{table}

We have evaluated our approach also on a pre-trained DeepLabv2 model from \citet{Chen2019DomainLoss} (with less powerful pre-training compared to \citep{Kundu2021GeneralizeSegmentation}). Our approach shows similar improvements compared to the pre-trained model also in this case -- see Table \ref{tab:sfda-msl} in the Appendix.

\section{Analysis}
\label{sec:analysis}
\begin{figure}[h!]
  \begin{center}
  \includegraphics[width=\columnwidth]{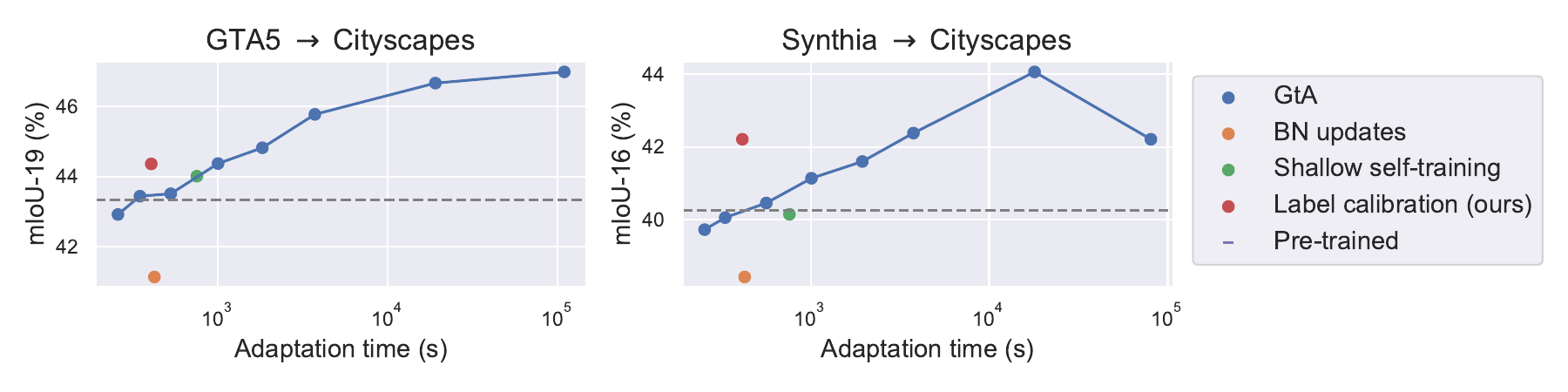}
  \caption{Comparison of the performance and adaptation time of the different methods.}
  \label{fig:adaptation_times}
  \end{center}
\end{figure}

We analyse the performance and adaptation time (when using a GPU) of the different methods in Figure~\ref{fig:adaptation_times} -- with \textit{GtA} trained for various numbers of iterations. The time required by our Label calibration method is similar to that of simply updating the BN statistics. However, standard back-propagation based SFDA method \citep{Kundu2021GeneralizeSegmentation} requires about 200-300x more time if run for the default number of iterations (right-most points in Figure \ref{fig:adaptation_times}). If given a similar amount of time as needed by our method, it gives only negligible improvements over the pre-trained model.

To provide further insights into our method we analyse the estimated prototypes. These are analysed in Figure \ref{fig:prototypes} in the Appendix and show that non-trivial soft-label prototypes are learned.

\section{Conclusion}
We have proposed a simple and efficient method to calibrate semantic segmentation labels under domain shift and obtain improvements in performance. As part of our method we calculate a soft-label prototype for each class under domain shift and make predictions by finding the nearest prototype. The method is orders of magnitude faster than existing back-propagation based unsupervised source-free domain adaptation methods, but it still leads to noticeable performance improvements.

\bibliographystyle{apalike}
\bibliography{references, more_refs}

\begin{thebibliography}{}

\bibitem[Bojarski et~al., 2016]{Bojarski2016EndCars}
Bojarski, M., Del~Testa, D., Dworakowski, D., Firner, B., Flepp, B., Goyal, P.,
  Jackel, L.~D., Monfort, M., Muller, U., Zhang, J., Zhang, X., Zhao, J., and
  Zieba, K. (2016).
\newblock {End to end learning for self-driving cars}.
\newblock In {\em arXiv}.

\bibitem[Chen et~al., 2017]{Chen2017DeepLab:CRFs}
Chen, L.-C., Papandreou, G., Kokkinos, I., Murphy, K., and Yuille, A.~L.
  (2017).
\newblock {DeepLab: Semantic image segmentation with deep convolutional nets,
  atrous convolution, and fully connected CRFs}.
\newblock In {\em TPAMI}.

\bibitem[Chen et~al., 2019]{Chen2019DomainLoss}
Chen, M., Xue, H., and Cai, D. (2019).
\newblock {Domain adaptation for semantic segmentation with maximum squares
  loss}.
\newblock In {\em ICCV}.

\bibitem[Cordts et~al., 2016]{Cordts2016TheUnderstanding}
Cordts, M., Omran, M., Ramos, S., Rehfeld, T., Enzweiler, M., Benenson, R.,
  Franke, U., Roth, S., and Schiele, B. (2016).
\newblock {The Cityscapes dataset for semantic urban scene understanding}.
\newblock In {\em CVPR}.

\bibitem[Csurka et~al., 2022]{csurka2022visual}
Csurka, G., Hospedales, T.~M., Salzmann, M., and Tommasi, T. (2022).
\newblock Visual domain adaptation in the deep learning era.
\newblock In {\em Synthesis Lectures on Computer Vision}.

\bibitem[Ganin et~al., 2016]{Ganin2016Domain-AdversarialNetworks}
Ganin, Y., Ustinova, E., Ajakan, H., Germain, P., Larochelle, H., Laviolette,
  F., Marchand, M., Lempitsky, V., Dogan, U., Kloft, M., Orabona, F., and
  Tommasi, T. (2016).
\newblock {Domain-adversarial training of neural networks}.
\newblock In {\em JMLR}.

\bibitem[Gulrajani and Lopez-Paz, 2020]{Gulrajani2020InGeneralization}
Gulrajani, I. and Lopez-Paz, D. (2020).
\newblock {In search of lost domain generalization}.
\newblock In {\em arXiv}.

\bibitem[He et~al., 2015]{He2015DeepRecognition}
He, K., Zhang, X., Ren, S., and Sun, J. (2015).
\newblock {Deep residual learning for image recognition}.
\newblock In {\em CVPR}.

\bibitem[Huang et~al., 2021]{Huang2021ModelData}
Huang, J., Guan, D., Xiao, A., and Lu, S. (2021).
\newblock {Model adaptation: historical contrastive learning for unsupervised
  domain adaptation without source data}.
\newblock In {\em NeurIPS}.

\bibitem[Jackson et~al., 2019]{Jackson2019StyleRandomization}
Jackson, P.~T., Atapour-Abarghouei, A., Bonner, S., Breckon, T., and Obara, B.
  (2019).
\newblock {Style augmentation: Data augmentation via style randomization}.
\newblock In {\em CVPR workshops}.

\bibitem[Jung et~al., 2020]{imgaug}
Jung, A.~B., Wada, K., Crall, J., Tanaka, S., Graving, J., Reinders, C., Yadav,
  S., Banerjee, J., Vecsei, G., Kraft, A., Rui, Z., Borovec, J., Vallentin, C.,
  Zhydenko, S., Pfeiffer, K., Cook, B., Fernández, I., De~Rainville, F.-M.,
  Weng, C.-H., Ayala-Acevedo, A., Meudec, R., Laporte, M., et~al. (2020).
\newblock {imgaug}.
\newblock \url{https://github.com/aleju/imgaug}.

\bibitem[Kouw and Loog, 2021]{kouw2021daReview}
Kouw, W.~M. and Loog, M. (2021).
\newblock A review of domain adaptation without target labels.
\newblock In {\em TPAMI}.

\bibitem[Kundu et~al., 2021]{Kundu2021GeneralizeSegmentation}
Kundu, J.~N., Kulkarni, A., Singh, A., Jampani, V., and Babu, R.~V. (2021).
\newblock {Generalize then adapt: source-free domain adaptive semantic
  segmentation}.
\newblock In {\em ICCV}.

\bibitem[Kundu et~al., 2020]{Kundu2020UniversalAdaptation}
Kundu, J.~N., Venkat, N., M, R., and Babu, R.~V. (2020).
\newblock {Universal source-free domain adaptation}.
\newblock In {\em CVPR}.

\bibitem[Li et~al., 2020]{Li2020ModelData}
Li, R., Jiao, Q., Cao, W., Wong, H.-S., and Wu, S. (2020).
\newblock {Model adaptation: unsupervised domain adaptation without source
  data}.
\newblock In {\em CVPR}.

\bibitem[Liang et~al., 2020]{Liang2020DoAdaptation}
Liang, J., Hu, D., and Feng, J. (2020).
\newblock {Do we really need to access the source data? Source hypothesis
  transfer for unsupervised domain adaptation}.
\newblock In {\em ICML}.

\bibitem[Michaelis et~al., 2019]{Michaelis2019BenchmarkingComing}
Michaelis, C., Mitzkus, B., Geirhos, R., Rusak, E., Bringmann, O., Ecker,
  A.~S., Bethge, M., and Brendel, W. (2019).
\newblock {Benchmarking robustness in object detection: Autonomous driving when
  winter is coming}.
\newblock In {\em arXiv}.

\bibitem[Richter et~al., 2016]{Richter2016PlayingGames}
Richter, S.~R., Vineet, V., Roth, S., and Koltun, V. (2016).
\newblock {Playing for data: Ground truth from computer games}.
\newblock In {\em ECCV}.

\bibitem[Ros et~al., 2016]{Ros2016TheScenes}
Ros, G., Sellart, L., Materzynska, J., Vazquez, D., and Lopez, A.~M. (2016).
\newblock {The SYNTHIA dataset: A large collection of synthetic images for
  semantic segmentation of urban scenes}.
\newblock In {\em CVPR}.

\bibitem[Saito et~al., 2018]{Saito2018MaximumAdaptation}
Saito, K., Watanabe, K., Ushiku, Y., and Harada, T. (2018).
\newblock {Maximum classifier discrepancy for unsupervised domain adaptation}.
\newblock In {\em CVPR}.

\bibitem[Schneider et~al., 2020]{Schneider2020ImprovingAdaptation}
Schneider, S., Rusak, E., Eck, L., Bringmann, O., Brendel, W., and Bethge, M.
  (2020).
\newblock {Improving robustness against common corruptions by covariate shift
  adaptation}.
\newblock In {\em NeurIPS}.

\bibitem[Sun and Saenko, 2016]{Sun2016DeepAdaptation}
Sun, B. and Saenko, K. (2016).
\newblock {Deep CORAL: Correlation alignment for deep domain adaptation}.
\newblock In {\em ECCV}.

\bibitem[Zhang et~al., 2021]{Zhang2021AdaptiveShift}
Zhang, M., Marklund, H., Dhawan, N., Gupta, A., Levine, S., and Finn, C.
  (2021).
\newblock {Adaptive risk minimization: learning to adapt to domain shift}.
\newblock In {\em NeurIPS}.

\end{thebibliography}

\appendix
\section{Appendix}

\begin{table}[h]
  \caption{Our approach works also when using an alternative pre-trained model. Mean mIoU (\%) and standard deviation across three runs. mIoU is reported across 19 classes for GTA5 $\rightarrow$ Cityscapes, and across 16 and 13 classes for Synthia $\rightarrow$ Cityscapes.}
  \label{tab:sfda-msl}
  \begin{center}
  \begin{tabular}{lcc}
    \toprule
    Approach     & GTA5 $\rightarrow$ Cityscapes & Synthia $\rightarrow$ Cityscapes \\
    \midrule
    Pre-trained model & 33.97 $\pm$ 0.00 & 28.17 $\pm$ 0.00 / 32.54 $\pm$ 0.00 \\
    Label calibration & 34.34 $\pm$ 0.01 & 29.58 $\pm$ 0.01 / 34.10 $\pm$ 0.01 \\
    \bottomrule
  \end{tabular}
  \end{center}
\end{table}

\begin{figure}[h!]
  \begin{center}
  \includegraphics[width=\columnwidth]{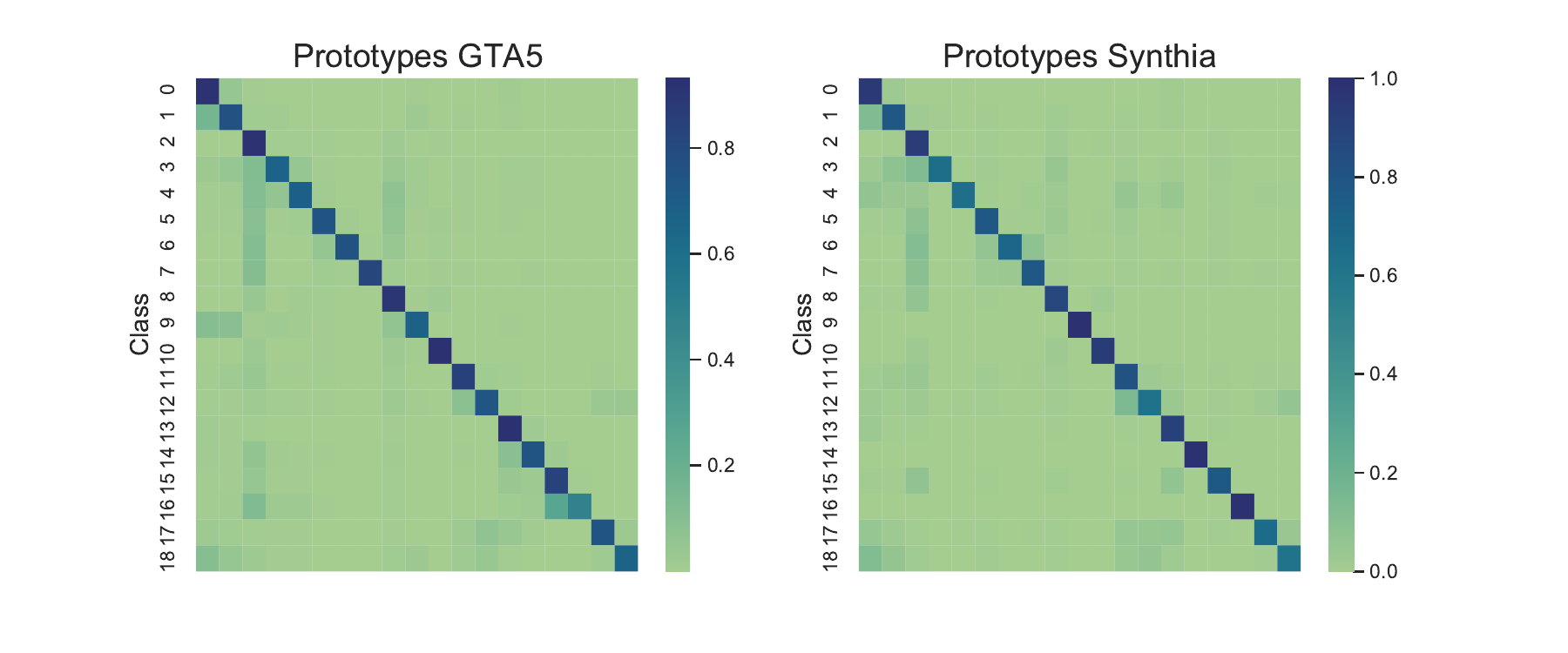}
  \caption{Illustration of the soft-label prototypes for the two different source datasets.}
  \label{fig:prototypes}
  \end{center}
\end{figure}

\end{document}